\documentclass[letterpaper, 10 pt, conference]{ieeeconf}  % Comment this line out if you need a4paper

\IEEEoverridecommandlockouts                              % This command is only needed if 
\usepackage{graphicx} % for pdf, bitmapped graphics files
\usepackage{subcaption}
\usepackage{amsmath}
\usepackage{amssymb}

\title{\LARGE \bf
QueryAdapter: Rapid Adaptation of Vision-Language Models in Response to Natural Language Queries
}

\author{Nicolas Harvey Chapman$^{1}$, Feras Dayoub$^{2}$, Will Browne$^{1}$ and Christopher Lehnert$^{1}$\vspace{-2em}%
\thanks{$^{1}$Nicolas Harvey Chapman, Will Browne and Christopher Lehnert are with the School of Electrical Engineering and Robotics, Queensland University of Technology, Brisbane, Australia ({\tt\small will.browne@qut.edu.au; c.lehnert@qut.edu.au; nicolasharvey.chapman@hdr.qut.edu.au}).}%
\thanks{$^{2} $Feras Dayoub is with the School of Computer Science and the Australian Institute of Machine Learning at the University of Adelaide, Adelaide, Australia ({\tt\small feras.dayoub@adelaide.edu.au}).}%
}

\begin{document}

\maketitle
\thispagestyle{empty}
\pagestyle{empty}

%%%%%%%%%%%%%%%%%%%%%%%%%%%%%%%%%%%%%%%%%%%%%%%%%%%%%%%%%%%%%%%%%%%%%%%%%%%%%%%%
\begin{abstract}

% This electronic document is a ÒliveÓ template. The various components of your paper [title, text, heads, etc.] are already defined on the style sheet, as illustrated by the portions given in this document.
A domain shift exists between the large-scale, internet data used to train a Vision-Language Model (VLM) and the raw image streams collected by a robot. Existing adaptation strategies require the definition of a closed-set of classes, which is impractical for a robot that must respond to diverse natural language queries. In response, we present QueryAdapter; a novel framework for rapidly adapting a pre-trained VLM in response to a natural language query. QueryAdapter leverages unlabelled data collected during previous deployments to align VLM features with semantic classes related to the query. By optimising learnable prompt tokens and actively selecting objects for training, an adapted model can be produced in a matter of minutes. We also explore how objects unrelated to the query should be dealt with when using real-world data for adaptation. In turn, we propose the use of object captions as negative class labels, helping to produce better calibrated confidence scores during adaptation. 
Extensive experiments on ScanNet++ demonstrate that QueryAdapter significantly enhances object retrieval performance compared to state-of-the-art unsupervised VLM adapters and 3D scene graph methods. Furthermore, the approach exhibits robust generalization to abstract affordance queries and other datasets, such as Ego4D.

\end{abstract}

%%%%%%%%%%%%%%%%%%%%%%%%%%%%%%%%%%%%%%%%%%%%%%%%%%%%%%%%%%%%%%%%%%%%%%%%%%%%%%%%
\section{INTRODUCTION}
% Foundational Vision-Language Models (VLMs) have enabled robots to detect \cite{clip, segment_anything, detic} and map objects \cite{clio, conceptgraphs, hovsg, bare, ovoslam} described using natural language. Such systems are not limited to a closed set of classes, and are thus able to generalize to diverse tasks and environments. To retain this generality, existing open-vocabulary robotic vision systems avoid adapting foundational VLMs using domain specific data \cite{conceptgraphs, clio}. However, an unintended outcome of this design is that the system may not be well-suited to the robot's current task or deployment environment \cite{seal, eal_semseg, self_improving, move_to_see, interactron}.  

Foundational Vision-Language Models (VLMs) have enabled robots to detect \cite{clip, segment_anything, detic} and map objects \cite{clio, conceptgraphs, hovsg, bare, ovoslam} described using natural language. Such systems are not limited to a closed set of classes, and are thus able to generalize to diverse tasks and environments.
However, a domain shift exists between the large-scale, internet data used to train a VLM and the raw image streams collected by a robot \cite{seal}. Consequently, pre-trained VLMs are unlikely to perform optimally in robotic deployment environments \cite{seal, eal_semseg, self_improving, move_to_see, interactron}.  

% Methods for adapting a VLM to domain specific data using few \cite{coop, clip_adapter} or no \cite{upl, ueo} labelled samples provide a natural solution to this problem. However, existing approaches require the definition of a closed-set of classes, which is impractical when the robot is expected to respond to diverse natural language queries. 
% As such, existing approaches to open-vocabulary object detection rely on pre-trained VLMs that underperform in robotic deployment environments \cite{conceptgraphs, clio}.

Methods for adapting a VLM to domain specific data using few \cite{coop, clip_adapter} or no \cite{upl, ueo} labelled samples provide a natural solution to this problem. However, existing approaches require the definition of a closed-set of classes to perform adaptation. In contrast, pre-trained VLMs are often used in robotics to respond to diverse natural language queries. In this setting, it is not sufficient to improve the performance of the VLM on a closed-set of classes. Instead, methods are required that enable an adapted model to be used for open-vocabulary object detection.

\begin{figure}[!t]
\centering
\includegraphics[width=1\columnwidth]{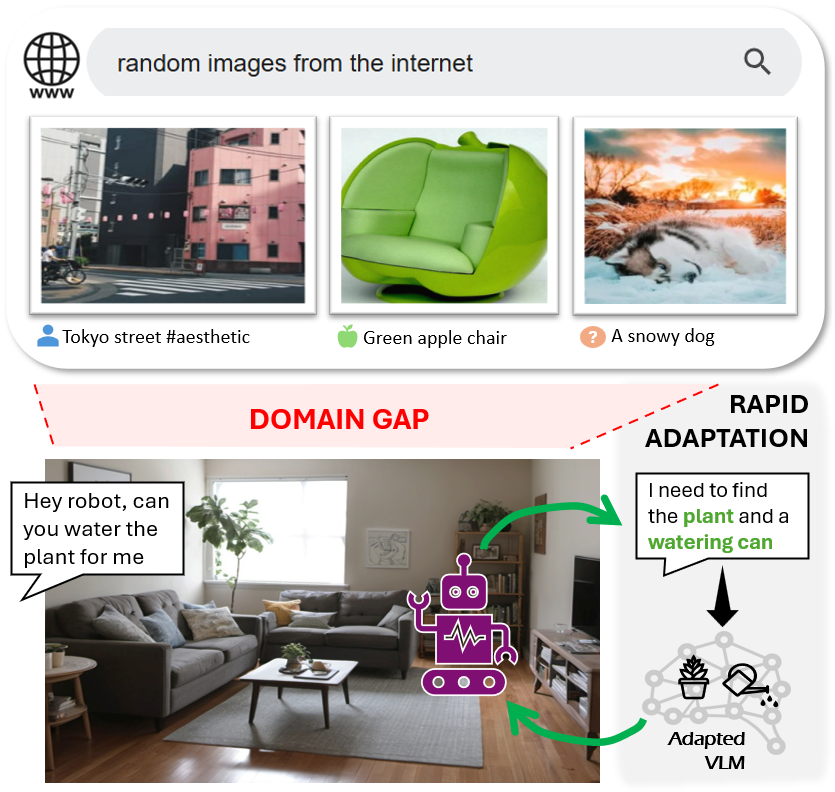}
\caption[Domain Shift for VLMs]{Existing methods for overcoming the domain gap between captioned images and robotic data streams require the definition of a closed-set of classes. This is unrealistic for robots that detect objects in response to diverse natural language queries. In response, we explore how a pre-trained VLM could be rapidly adapted to natural language queries as they arise. 
This approach avoids having to pre-define a closed-set of classes, ensuring that an adapted model can be used for open-vocabulary object detection.
% An example of the captioned image data used to train a VLM (top) \cite{laion5b}. Such datasets contain images that are not representative of robotic deployment, where diverse tasks must be performed using an egocentric data stream.
}
\vspace{-1.5em}
\label{domain_shift}
\end{figure}

% Thus, to improve open-vocabulary object detection in robotic deployment environments, we explore how a pre-trained VLM can be adapted \textit{without} pre-defining a closed-set of classes.
To overcome this limitation, we explore how a pre-trained VLM (e.g. CLIP) can be adapted for use in robotics \textit{without} pre-defining a closed-set of classes. 
We aim to leverage the fact that existing robotic vision systems \cite{conceptgraphs, clio, hovsg} perform open-vocabulary object detection in response to natural language queries (e.g. `water the plant'). Furthermore, using parameter-efficient methods such as prompt tuning \cite{coop, cocoop}, adaptation of a VLM can be performed without fine-tuning the entire model. Thus, we hypothesise that a pre-trained VLM could be quickly adapted to natural language queries as they arise (Figure \ref{domain_shift}). This avoids having to pre-define a closed-set of classes, ensuring that an adapted model can be used for open-vocabulary object detection.

% This way, an improved model could be used to carry out the task, without needing to pre-define a closed-set of classes. 

% We hypothesise that a VLM could be quickly adapted to an open-vocabulary tasks description (e.g. water the plant). In turn, the robot could use the adapted model to carry out the task, before repeating the process as new tasks arise. 
% This way, an adapted model can be applied to open-vocabulary detection tasks that were previously unknown to the robot.

% To overcome this limitation, we explore how a pre-trained VLM can be adapted \textit{without} pre-defining a closed-set of classes. To achieve this, a robotic vision paradigm is proposed where the VLM (e.g. CLIP) is quickly adapted to open-vocabulary detection tasks as they arise (Figure \ref{hook_figure}). 
% % The goal of this approach is to retain the generality of using a pre-trained VLM \cite{clip, conceptgraphs}, while ensuring that an optimal model is applied to specific tasks \cite{seal}. 
% Given a new task related to a small set of objects (e.g. water the plant), we leverage unlabelled data collected by the robot in previous deployments to efficiently adapt the model. The robot can then use the adapted model to carry out the task, before repeating the process as new tasks arise. 
% This way, an optimised model can be used to respond to diverse natural language queries.

\begin{figure}[!t]
\centering
\includegraphics[width=1\columnwidth]{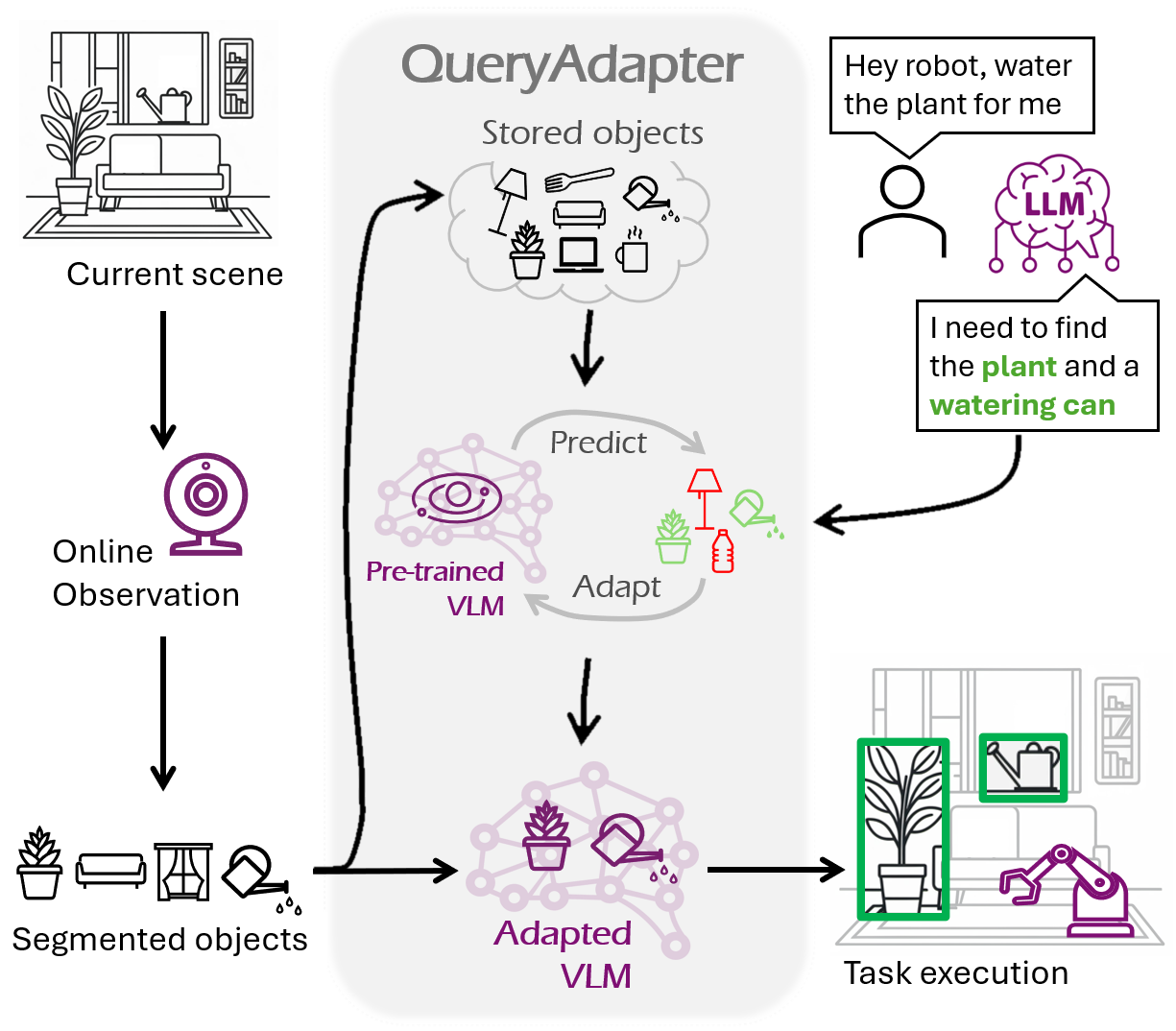}
\caption{Our proposed framework for rapidly adapting a pre-trained VLM to detect objects relevant to a natural language query. Given a new query, we use an LLM to generate a set of ``target classes'' required to fulfill the request. Unlabelled data collected by the robot in previous deployments is then used to align VLM features with these target classes. As a final step, the adapted model is used to detect the target classes in the current scene.
}
\vspace{-1.5em}
\label{simple_method}
\end{figure}

To this end, we propose QueryAdapter; a novel robotic vision framework for rapidly adapting a pre-trained VLM in response to a natural language query (Figure \ref{simple_method}). Given a new query, we use a Large Language Model (LLM) to generate a set of ``target classes'' required to fulfill the request. Unlabelled data collected by the robot in previous deployments is then used to align VLM features with these target classes. To improve efficiency, only the top $k$ previously observed objects for each target class are selected for adaptation. Similarly, learnable prompt tokens \cite{coop} are optimised instead of fine-tuning the entire model, allowing adaptation to be performed in a few minutes. 
As a final step, the adapted model is used to detect the target classes in the current scene, improving the retrieval of objects related to the query. 

% As a final step, the adapted model and the target classes are used to retrieve objects from the current scene that are relevant to the task. In this way, an improved model can be used to complete novel tasks without pre-defining a closed-set of relevant classes.

% The approach firstly uses an LLM to process a natural language task description, generating a set of  required to complete the task.

% Using this framework (Figure \ref{hook_figure}) an improved model can be applied to open-vocabulary detection tasks without needing to pre-define a closet-set of relevant classes

% This framework requires that adaptation be performed quickly to minimise downtime of the robot. Furthermore, as the tasks are unknown before deployment, unlabelled data collected in previous environments must be leveraged to perform adaptation.

% When implementing QueryAdapter, we find that the majority of previously observed objects are irrelevant to a given natural language query. 
A further challenge in implementing this framework is that for a specific query, very few objects will be relevant. In existing literature, objects that fall outside the classes defined for detection are termed Out-of-Distribution (OOD) \cite{ueo} or open-set objects \cite{open_vlm}. A similar problem occurs when implementing QueryAdapter, with the majority of previously observed objects being unrelated to a given natural language query. 
% Existing unsupervised learning methods perform poorly in the presence of such \textit{open-query} objects.
We refer to these as \textit{open-query} objects, and find that existing unsupervised learning methods perform poorly in this challenging setting.
% In this realistic setting, we find that existing unsupervised VLM adapters designed for image classification perform poorly \cite{ueo, upl}.
% This produces an extremely challenging OOD problem that is akin to finding a needle in a haystack \cite{ueo, open_vlm, probvlm}. 
% This produces a challenging Out-of-Distribution (OOD) problem in which existing unsupervised learning methods perform poorly \cite{ueo, upl}. 
% To overcome this, we propose a novel approach to performing adaptation using an unlabelled data stream containing many OOD objects. 
% To overcome this, we propose using the captions of previously observed objects to extract the most common classes in the dataset, and propose these as ``negative classes''. 
To overcome this, we use the captions of previously observed objects to extract the most common classes in the dataset, and propose using these as ``negative classes'' during adaptation.
The addition of these negative classes helps produce better calibrated confidence scores, improving the performance of unsupervised learning techniques based on entropy maximisation \cite{ueo}. This method ensures that the QueryAdapter framework remains effective when using data collected by the robot in previous deployments.

To summarise, our work makes the following contributions: 
\begin{itemize}
    \item We propose QueryAdapter, a framework for rapidly adapting a pre-trained VLM in response to a natural language query.
    This avoids having to pre-define a closed-set of classes, ensuring that an adapted model can be used for open-vocabulary object detection.
    % This is the first approach for adapting a VLM in response to a specific query, avoiding the need to pre-define a closed-set of classes.
    % responding to task-oriented queries with an adapted VLM. Using this approach, an improved model can be used to complete novel tasks without pre-defining a closed-set of relevant classes.
    % \item A novel pseudo-labelling approach that produces a dataset for adapting a VLM to a task described using natural language.
    \item We propose the use of object captions as negative classes during adaptation, helping to produce better calibrated confidence scores for open-query objects. This ensures that QueryAdapter is effective when using the raw data collected by the robot in previous deployments. 
    \item We conduct a detailed evaluation of natural language object retrieval in real-world scenes \cite{scannet++}, demonstrating the effectiveness of our approach in adapting to challenging natural language queries. 
    % \item An approach to align object features with a core set of natural language concepts using only unlabelled data collected by the robot, leading to improved retrieval of core concepts without sacrificing zero-shot performance. 
    % \item A novel pseudo-labelling approach that can assign any number of core concepts to a previously observed object, effectively dealing with many out-of-distribution samples and diverse descriptions of the same object. 
    % \item A demonstration of the system for continual robot learning, where core-concepts and unlabelled data are incrementally made available to the robot. 
\end{itemize}

\section{RELATED WORK}
\subsection{Language-Image Pre-training}
The availability of captioned images has allowed the joint training of image and text encoders using natural language supervision \cite{clip}. The seminal work in this space is Contrastive Language-Image Pretraining (CLIP), which uses a contrastive loss to produce similar embeddings for image-text pairs \cite{clip}. The resulting VLM can be used to perform image classification with open-vocabulary classes \cite{clip}, and exhibits superior resistance to domain shift compared with supervised pre-training methods \cite{clip, grounding_with_text}. 
% A variety of methods extend have extended this capability to perform dense prediction tasks such as object detection \cite{glip, owl, detic} and semantic segmentation \cite{lseg, dense_clip, ZegCLIP, segment_anything}. 
However, a domain shift exists between the large-scale, internet data used to train these VLMs and the raw image streams collected by a robot \cite{seal} (Figure \ref{domain_shift}). Consequently, a pre-trained VLM is unlikely to perform optimally in robotic deployment environments.

\subsection{Object Retrieval from Natural Language}
A range of robotic vision systems have been proposed for responding to natural language queries in a real-world scenes \cite{clio, conceptgraphs, hovsg, bare, ovoslam}. These systems are distinct from traditional object detection and mapping systems, as they do not aim to classify each object at the time of observation \cite{kimera, multi-tsdfs}. Instead, they look to maintain a generic representation of the environment that can be utilised later to respond to natural language queries \cite{vlmap, lerf}. To solve this problem, foundational VLMs are used to produce objects segmentations \cite{segment_anything} and open-vocabulary features \cite{clip, detic, glip, owl, lseg} from a stream of posed RGB-D images. While these segments can be directly used to respond to queries \cite{bare}, they are often fused across frames using projective geometry and feature similarity to produce distinct object instances \cite{conceptgraphs, hovsg, clio, ovoslam}. In turn, relationships between instances can be predicted to generate a complete 3D Scene Graph (3DSG) \cite{conceptgraphs, hovsg, ovoslam}. 

To respond to natural language queries, the cosine similarity between text and object embeddings can be used to retrieve relevant objects. Alternatively, Multi-modal Large-Language Models (MLLMs) \cite{llava, llava3d} can generate object captions and select those that are relevant to a particular query \cite{conceptgraphs}. What is common across these systems is that they rely on pre-trained vision-language models that have not been adapted for use in robotic deployment environments. In response, we propose an approach to adapt a VLM to a particular natural language query, improving the retrieval of relevant objects.

\subsection{Adaptive Embodied Object Detection}
Several works aim to close the domain gap between the large-scale, internet data used to train a VLM and the raw image streams collected by a robot \cite{seal}. Embodied Active Learning (EAL) methods use the spatial-temporal consistency of a scene as a learning signal to perform adaptation \cite{seal, eal_semseg, self_improving, move_to_see}. However, existing EAL methods focus on using closed-vocabulary object detectors. Instead, we aim to use previously observed objects to quickly adapt a VLM to open-vocabulary concepts. Attempts have also been made to optimise the fusion of CLIP features across different views of the scene \cite{ovoslam, bare, eod}. Unlike these systems, we rely solely on unlabelled data collected by the robot to perform adaptation, and do not require the definition of a closed-set of classes. 

% \begin{figure}[h]
% \centering
% \includegraphics[width=0.56\textwidth]{Figures/Chapter-2/vlm_domain_shift.png}
% \caption[Domain Shift for VLMs]{\label{fig:vlm_domain_shift} An example of the captioned image data from the LAION-5B dataset (top). Such datasets contain images that are not representative of robotic deployment, where diverse tasks must be performed using the data stream collected by a robot. Image adapted from \cite{laion5b}.}
% \end{figure}

\subsection{Parameter-efficient Transfer Learning of VLMs}
Recent work has focussed on adapting a VLM to particular downstream tasks without altering the pre-trained image and text encoders \cite{clip_adapter, coop, cocoop, multi_modal_adapter, vpt, tip_adapter}. This allows adaptation to be performed with limited data, without damaging the representations learnt during large-scale pre-training. The most basic approach, fitting a linear probe on top of the image encoder, was explored in the original work on CLIP \cite{clip}. Prompt tuning \cite{coop} formalised the task of parameter-efficient adaptation of VLMs and proposed learning a set of context embeddings that can be prepended to the tokenised class names to enhance image classification performance. Various other adapters have since been proposed to augment the text features \cite{coop, cocoop, tip_adapter} and image features \cite{clip_adapter, vpt} using limited labelled data. 

Motivated by the Unsupervised Domain Adaptation (UDA) literature \cite{domain_theory, uda_survey, udaod1, udaod2}, attempts have been made to adapt CLIP using only unlabelled data \cite{ueo, upl}. Unsupervised Prompt Learning (UPL) \cite{upl} selects the top $k$ confident samples for each class in the unlabelled data, and uses them as pseudo-labels to fine-tune the system via prompt tuning \cite{coop}. Universal Entropy Optimisation (UEO) \cite{ueo} extends the standard unsupervised learning task to consider the existence of OOD images in the training data. They propose a learning objective that minimises entropy for confident samples while maximising it for low-confidence samples, and use this to perform prompt tuning. However, both UEO and UPL require the definition of a closed-set of classes, which is undesirable when the robot is expected to respond to diverse natural language queries. Furthermore, we find that these approaches fail when applied to a robotic data stream with many open-query objects.  

% While promising, we find that this method performs poorly for complex natural language queries and in the presence of many OOD samples. Universal Entropy Optimisation (UEO) extends the standard unsupervised learning task to consider the existance of OOD images in the training data. They propose a learning objective that minimises entropy for confident samples while maximising it for low-confidence samples, and use this to perform prompt tuning. However, as it is designed for image classification, entropy minimisation assumes that each object has only one correct label. Furthermore, we find that this approach degrades significantly when facing a significant number of OOD samples. In response to these limitations, we propose a novel pseudo-labelling approach designed specifically to deal with these challenges. Similar to UPL, we select the top-k confident objects for each core concept as candidate pseudo-labels. We then pass these candidate objects to a MLLM to produce a caption and decide if the candidate pseudo-label is likely to be correct. This way, we significantly reduce the number of OOD samples in our training data. We then use these pseudo-labels to train an adapter to perform object retrieval, significantly improving performance on the core concepts without compromising zero-shot performance.

\subsection{OOD and Open-set Detection with VLMs}
VLMs are trained to classify open-vocabulary concepts, making them robust to open-set conditions. However, the act of defining a query set introduces closed-set assumptions, in turn making VLMs vulnerable to open-set \cite{open_vlm, bare} or OOD \cite{ueo} objects. Recent work attempts to overcome this by using random words and embeddings as negative classes \cite{open_vlm}, and by adapting CLIP to better express prediction uncertainty \cite{probvlm}. 
In this work, we study a specialised version of this problem where objects unrelated to a specific query (termed open-query objects) must be rejected during adaptation.
% In turn, we extend existing work in unsupervised VLM adaptation \cite{ueo} to deal with such objects, which we term open-query objects. 
In response, we generate captions for all objects in the training data and use the most common class names as negative classes. As a further step to filter our open-query objects and improve learning efficiency, we only the use the top $k$ previously observed objects for adaptation.

\section{Preliminaries}

\begin{figure}[!t]
\centering
\includegraphics[width=1\columnwidth]{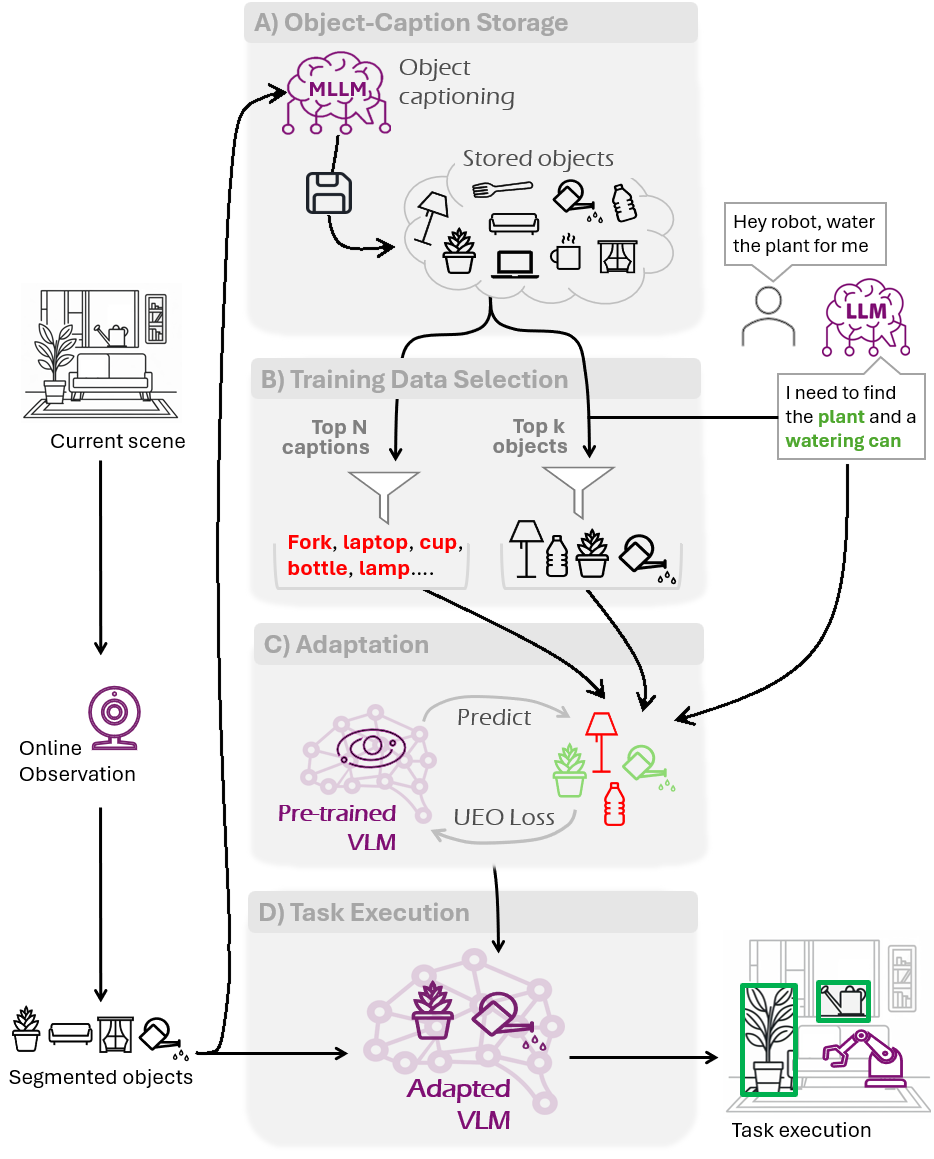}
\caption{A detailed summary of QueryAdapter, the proposed framework for responding to natural language queries with an adapted VLM. The method is split into four steps; object captioning and storage, training data selection, adaptation and object retrieval.
% The first input to the system is a task command from the user (for example, water the plant for me). An LLM is then used to decompose the request into a set of target classes required to complete the task. These classes are passed to our TaskAdapter module, which leverages unlabelled objects from previous deployments to quickly optimise the pre-trained VLM for detecting the target objects. The robot then uses the adapted VLM in its online perception system to retrieve objects from the current scene that are relevant to the task. Upon execution of the task, the stored objects can be updated with the object segments from the current scene and the process repeated when a new task arises.
% More specifically, stored objects from previous deployments are used in a novel self-training approach to optimise a light-weight adapter network. 
}
\vspace{-1.5em}
\label{method_figure}
\end{figure}

\subsection{Problem Formulation}
\label{prob_form}
We consider the scenario where a robot collects from the current scene $j$ a sequence of posed RGB-D images. The following formulation can also be applied to a single posed image without losing generality. As per ConceptGraphs \cite{conceptgraphs}, pixel segmentation masks $m_{x}$ are firstly extracted for each image. Image crops $c_{x}$ for each segment are then passed to an image encoder $g_{I}\left(\cdot \right)$ to produce an open-vocabulary feature $I_{x}$. This image encoder has a corresponding text encoder $g_{T}\left(\cdot \right)$ that can produce text features $T_{q}$ in the same embedding space as $I_{x}$. Depth, pose and the camera intrinsics are then used to project each pixel mask into the world frame, generating a point cloud $p_{x}$. The result of this process is that for the current scene, we obtain a set of $X$ object segments each defined by a segmentation mask $m_{x}$, image crop $c_{x}$, open-vocabulary image feature $I_{x}$ and point cloud $p_{x}$:
\begin{equation}
    O_{j} = \left\{\left( m_{x}, c_{x}, I_{x}, p_{x} \right)\right\}^{X}_{x=1}
\end{equation}
% \textit{Object Retrieval from Natural Language: }
Given a natural language query $Q$, the robot must retrieve a relevant object from the set $O_{j}$. The retrieved object can then be localised within the scene using the pre-computed point cloud $p_{x}$. We further assess a more complex version of this task where the robot is given a natural language task description $Q_{t}$ that it must complete in the environment. In response to such a query, the robot must return a set of objects from $O_{j}$ that are required to complete the task.

\subsection{Object Retrieval with Using Cosine Similarity}
\label{sec:retrieval}
To retrieve objects related to a natural language query, the query $Q$ can be passed to the text encoder $g_{T}\left(\cdot \right)$ to produce a text feature $T_q$:
\begin{equation}
    T_q = g_{T}\left(Q \right)
    \label{eq:text_encoder}
\end{equation}
The similarity $s_x$ between an object feature $I_x$ and the text feature $T_q$ can then be calculated as:
\begin{equation}
    s_x = \mathbb{S}(I_x, T_q)
    \label{eq:cosine_sim}
\end{equation}
where $\mathbb{S}(\cdot, \cdot)$  denotes cosine similarity. The object segment $x$ with the highest similarity $s_x$ can then be returned in response to the query. Alternatively, the top $k$ segments can be returned as required.

% \subsection{Object Retrieval with Using LLMs}
% Other works use an LLM but 1) merging segments to create distinct instances 2) captioning each instance using a MLLM and 3) using the MLLM

\section{QueryAdapter Method}
\label{sec:adaptation}
Next, we define our QueryAdapter framework for adapting the pre-trained VLM in response to a natural language query $Q$. We separate this into four steps that are described in the following sections and summarised in Figure \ref{method_figure}.

\subsection{Object Captioning and Storage}
Our proposed robotic vision system relies on a set of previously observed objects to perform adaptation. Furthermore, captions are required for these objects to produce negative labels for improving adaptation in the presence of open-query objects. Upon completion of deployment in scene $j$, a caption $\hat{c}_{x}$ is produced for each object segment in $O_{j}$ by passing the extracted image crops to a MLLM.
\begin{equation}
    \hat{c}_{x} = MLLM\left( c_{x}\right)
\end{equation}
These captions are added to the set of object segments, and the image crops removed for efficient storage. Furthermore, we add the index for the current scene. The set of object segments in the scene thus becomes:
\begin{equation}
    O_{j} = \left\{\left(j, m_{x}, \hat{c}_{x}, I_{x}, p_{x} \right)\right\}^{X}_{x=1}
\end{equation}
Note that the use of a MLLM in this process is computationally expensive, which is why we perform this operation offline, after deployment. Lastly, the updated set of objects $O_{j}$ are added to the current set of stored object $S_{j}$. This update rule is defined as:
\begin{equation}
    S_{j+1} = S_{j}+O_{j}
\end{equation}
In our experiments, we do not iteratively deploy the robot across many scenes to generate the stored objects. Instead, we produce $S_{j}$ using a predefined set of $j$ scenes.

\begin{figure*}[t!]
\quad
    \begin{subfigure}[t]{0.3\textwidth}
    \centering
    \includegraphics[trim={0.3cm 0 -0.5cm 0},clip,width=6cm]{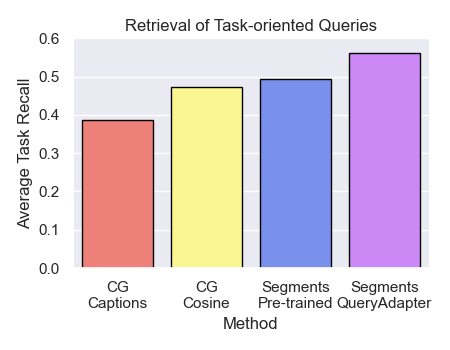}
    \caption{Comparison of QueryAdapter with methods based on 3DSGs for task-oriented object retrieval.}
    \end{subfigure}
\quad
    \begin{subfigure}[t]{0.3\textwidth}
    \centering
    \includegraphics[trim={0.3cm 0 -0.5cm 0},clip,width=6cm]{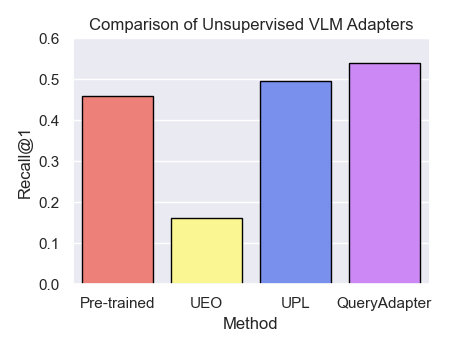}
    \caption{Comparison of QueryAdapter with unsupervised VLM adapters using small sets of target classes.}
    \end{subfigure}
\quad
    \begin{subfigure}[t]{0.3\textwidth}
    \centering
    \includegraphics[trim={0.3cm 0 -0.5cm 0},clip,width=6cm]{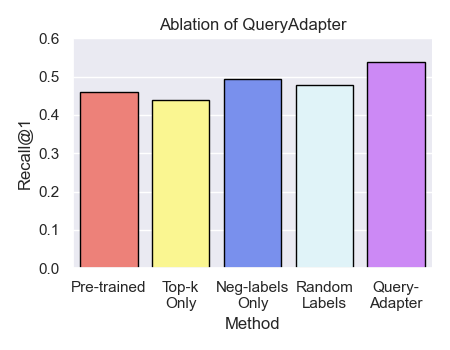}
    \caption{Ablation of QueryAdatpter using small sets of target classes.}
    \end{subfigure}
\caption{Comparison of QueryAdapter with state-of-the-art unsupervised VLM adapters and 3DSG methods.}
\label{comparison_results}
\vspace{-1.5em}
\end{figure*}

\subsection{Training Data Selection}
\label{sec:training_data}
The next step aims to use the stored objects $S_{j}$ to create training data for adapting the VLM to the query $Q$. Firstly, a LLM is used to decompose the query into a set of target classes $\left\{C_{t} \right\}_{t=1}^{T}$ required to fulfil the request:
\begin{equation}
    \left\{C_{t} \right\}_{t=1}^{T} = LLM\left( Q\right)
\end{equation}
% Alternatively, the target classes $C_{t}$ could be directly defined by the user. However, our method demonstrates that more complex adaptation behaviour is possible using the LLM.   
Then, we use the caption $\hat{c}_{x}$ for each object in $S_{j}$ to produce a set of negative classes $\left\{C_{n} \right\}_{n=1}^{N}$.
The nouns are extracted from each caption, and the most common $N$ nouns used to define the set of negative labels. 
To avoid overlap between the negative and target classes, an LLM is used to remove negative classes that have a direct synonym in the target classes. 
The concatenation of the resulting negative and target classes defines the total set of classes $\left\{C_{a} \right\}_{a=1}^{A}$ used to adapt the model:
\begin{equation}
    \left\{C_{a} \right\}_{a=1}^{A} = \left\{C_{t} \right\}_{t=1}^{T} + \left\{C_{n} \right\}_{n=1}^{N}
\end{equation}
Lastly, we select a subset of the stored objects as samples for performing adaptation. This is done to reduce training time and focus adaptation towards the target classes.
For each scene $j$, the top $k$ most similar object segments are retrieved for each target class as described in Section \ref{sec:retrieval}.
% We do this per scene to ensure that the same process can be applied regardless of the size of the stored objects. 
The retrieved object segments, which are analogous to pseudo-labels, are then added to the set of filtered objects to be used for adaptation:
\begin{equation}
    F_{j} = topk\left(S_{j}, \left\{C_{t} \right\}_{t=1}^{T} \right), F_{j} \subset S_{j}
\end{equation}

\subsection{Adaptation}
Next, we use the filtered set of object segments $F_{j}$ and the adaptation classes $\left\{C_{a} \right\}_{a=1}^{A}$ to adapt the VLM. Following CoOp \cite{coop}, we freeze the text and image encoders and instead optimise the text prompts used to perform classification. Specifically, we define a set of $m$ learnable word vectors $\left\{\left[V_{i} \right]\right\}_{i=1}^{m}$ to generate a prompt of the form ``[V$_1$], [V$_2$], \dots, [V$_m$], [CLASS]'' for each adaptation class. As per Eq. (\ref{eq:text_encoder}), the learnable prompts for each adaptation class can be passed to the text encoder to generate corresponding text features $\left\{T_{a} \right\}_{a=1}^{A}$. The probability that an object segment with embedding $I_{x}$ belongs to class $a$ can then be defined using the softmax operation:
\begin{equation} 
\label{eq:clip}
p_a(x) = \frac{\exp(\mathbb{S}(I_x, T_a)/ \tau)} {\sum_{i=1}^{A} \exp(\mathbb{S}(I_x, T_i)/\tau)},
\end{equation}
where $\mathbb{S}(\cdot, \cdot)$ denotes the cosine similarity and $\tau$ is the temperature parameter. If using labelled data to perform adaptation, a standard image classification loss could in turn be applied to optimise $\left\{\left[V_{i} \right]\right\}_{i=1}^{m}$. However, to leverage the unlabelled object segments in $F_{j}$, we implement the unsupervised loss proposed by UEO \cite{ueo}. This loss aims to minimise entropy for samples with a confident classification while maximising it for uncertain samples. To achieve this, the maximum softmax probability score returned via Eq. (\ref{eq:clip}) is used as an estimate of confidence, denoted $w\left(x\right)$. The following loss is then used to achieve simultaneous entropy minimisation and maximisation:
\begin{equation}
\mathcal{L} = \sum_{x \in \mathcal{B}_t} \widetilde{w}(x) \mathcal{H}(p(x)) - \mathcal{H}(\bar{p}),\; 
\label{eq:went2}
\end{equation}
where $\mathcal{B}_t$ is a training mini-batch sampled from $F_{j}$, $\widetilde{w}(x)$ is the normalised value of $w(x)$ across the mini-batch and $\mathcal{H}(\cdot)$ is the Shannon entropy of a probability distribution. Furthermore, $\bar{p}$ is the inversely weighted average of predictions for each sample within the mini-batch:
\begin{equation}
\bar{p} = \sum_{x \in \mathcal{B}_t} \frac{p(x)}{\widetilde{w}(x)}
\end{equation}
We find that when relatively few classes $\left\{C_{a} \right\}_{a=1}^{A}$ are defined for adaptation, the estimation of confidence $w(x)$ and entropy $\mathcal{H}(p(x))$ are much less reliable. By utilising additional negative classes $\left\{C_{n} \right\}_{n=1}^{N}$ during adaptation, we aim to obtain a better prediction for these values. Furthermore, by selecting only those samples similar to the target classes $\left\{C_{t} \right\}_{t=1}^{T}$, we aim to reduce the complexity of the open-query problem, allowing the entropy maximisation term of Eq. (\ref{eq:went2}) to dominate.

\subsection{Object Retrieval}
Once adaptation is performed, which needs to occur quickly, the optimized prompt vectors $\left\{\left[V_{i} \right]\right\}_{i=1}^{m}$ can be used to retrieve object segments from the scene. We prepend these vectors to the target classes $\left\{C_{t} \right\}_{t=1}^{T}$ to generate the optimised prompts $\left\{P_{t} \right\}_{t=1}^{T}$. We can then use these prompts to retrieve object segments relevant to the natural language query, as described in Section \ref{sec:retrieval}.

% Passing these prompts through the text encoder as per  Eq. (\ref{eq:text_encoder}), we get text features $\left\{T_{t} \right\}_{t=1}^{T}$. Finally, to retrieve object segments relevant to the task description, we simply apply Eq. (\ref{eq:cosine_sim}) for each text feature $T_{t}$.

\section{RESULTS}
\subsection{Experimental Settings}

% \begin{figure}[!t]
% \centering
% \includegraphics[width=1\columnwidth]{Figures/task_queries_bar_charts.png}
% \caption{
% Comparison of the proposed robotic vision paradigm with ConceptGraphs for task-oriented object retrieval in the Scannet++ dataset.
% }
% \label{task_oriented}
% \end{figure}

\textbf{Dataset Preparation:} Following similar work \cite{bare, ovoslam}, we utilise scenes from the Scannet++ dataset \cite{scannet++} for both adaptation and evaluation. The standard split that assigns 230 scenes to training and 50 scenes to testing is used. We pre-process the raw data for each scene to produce the set of object segments $O_{j}$ as described in Section~\ref{prob_form}. We use SegmentAnything \cite{segment_anything} to produce segmentation masks for each image and CLIP \cite{clip} to generate the open-vocabulary embeddings. The number of scenes $j$ used for adaptation is by default defined as 70, however we test the sensitivity of this variable in later experiments.

% \textbf{Evaluation of Task-oriented Adaptation: }We utilise two distinct experimental approaches to assess task-oriented adaptation using Scannet++ \cite{scannet++}. Firstly, we define a set of natural language task descriptions and use the top-100 most common classes to label the relevant objects required to complete the task. 
% % In response to a task description, we use the segments from $n$ training scenes to apply TaskAdapter as per Section III. 
% The robot can then be evaluated on its ability to retrieve objects that are in the labelled set of relevant objects. This experiment results in 158 task queries relating to 329 relevant objects for evaluating our overall robotic vision pipeline. These annotated tasks will be made available on our github repository.

\textbf{Evaluation of Task-oriented Object Retrieval:}
To assess object retrieval in response to complex natural language queries, we define a set of task descriptions that can be completed in the Scannet++ test scenes \cite{scannet++}. In turn, the ``relevant classes'' required to complete each task are defined using the top-100 common classes in the dataset. In scenes where all relevant classes are present, the robot is evaluated on its ability to retrieve these objects in response to the query. This experiment results in 158 queries relating to 329 relevant objects for evaluating QueryAdapter. \footnote{The annotated queries and code for performing evaluation will be made available on our github repository on acceptance.\label{myfootnote}}

% Given a task description, we use the object segments from $j$ training scenes to apply TaskAdapter as per Section III. We then perform object retrieval using the adapted model, making sure to only evaluate a task in scenes where a correct solution is possible. 
To evaluate segments retrieved by the robot, we use the ground truth point cloud to assign a class label to each segment. We firstly assign each point in the object segment the label of the closest ground-truth point, before assigning the most common label across all points to the segment. 
% In addition, if more than 5 objects are retrieved in response to a query, we randomly sample 5 to preform evaluation.
Using these ground truth labels, we can calculate the proportion of relevant classes that were recalled in response to the query. We report the average of this metric across all tasks as the Average Task Recall (ATR).

\textbf{Optimisation Procedure:}
To avoid biasing the method towards the queries used for evaluation, we optimise QueryAdapter using a different experimental approach. To simulate adapting to natural language queries, we randomly generate small sets of target classes that the robot must adapt to. We use the most common object classes in Scannet++ to define eight sets of six target classes. These target classes can then be used to perform adaptation as per Section~\ref{sec:adaptation}, with the query decomposition step skipped. 
% Following the example of existing work, we also generate a second set of pseudo-tasks using affordance queries \cite{conceptgraphs}. These are generated by asking an LLM to define the most common use case for each class. 
We report the recall@1 averaged across all target classes, which comprises a significant number of object queries ($>$1500). This allows us to optimise QueryAdapter without overfitting to particular types of natural language query. Furthermore, we use this dataset to compare with existing unsupervised VLM adapters.

\textbf{Baseline methods: }We firstly compare QueryAdapter with other unsupervised VLM adapters from the literature. UPL \cite{upl} uses top $k$ pseudo-labelling and cross-entropy loss to perform prompt learning. UEO \cite{ueo} uses the same self-training loss as QueryAdapter, but without negative labels or top $k$ filtering to deal with the many OOD objects in the unlabelled data stream. We also compare with an approach to object retrieval from natural language based on 3DSGs. We implement ConceptGraphs \cite{conceptgraphs} and perform object retrieval using both the object captions and cosine similarity.

\textbf{Implementation details: }For all experiments, we use the \textit{ViT-H-14} CLIP model from Openclip \cite{clip} as the pre-trained VLM. For the prompt learner \cite{coop}, we use four context vectors initialised with the prompt ``a photo of a'' and the temperature parameter $\tau$ is set to 0.01. We train the adapters with the Adam optimiser for 50 epochs on a single A100 GPU, with a batch size of 256 and learning rate of 0.0005. For QueryAdapter, we use the optimal setting of $k=8$ and 100 negative queries as standard. For UPL, we also find that $k=8$ is optimal. The number of negative labels is set as $N=100$. We use \textit{Llama-3-8B-Instruc} model as the LLM and \textit{llava-v1.6-vicuna-7b} as the captioning system for QueryAdapter and ConceptGraphs.\footnote{Examples of the prompts for both these models will be made available on our github repository on acceptance.\label{myfootnote}}

% \begin{figure}[!t]
% \centering
% \includegraphics[width=1\columnwidth]{Figures/ablation_bar_charts.png}
% \caption{
% Ablation of TaskAdatpter for class and affordance queries in the Scannet++ dataset.
% }
% \label{ablation}
% \end{figure}

\begin{figure*}[t!]
\vspace{-1em}
\quad
    \begin{subfigure}[t]{0.3\textwidth}
    \centering
    \includegraphics[trim={0.3cm 0.5cm -0.5cm 0},clip,width=6cm]{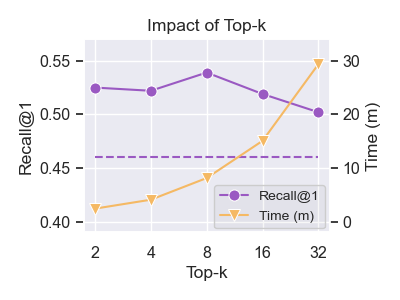}
    \caption{Impact of top $k$ setting on QueryAdapter performance and training time.}
    \end{subfigure}
\quad
    \begin{subfigure}[t]{0.3\textwidth}
    \centering
    \includegraphics[trim={0.3cm 0.5cm -0.5cm 0},clip,width=6cm]{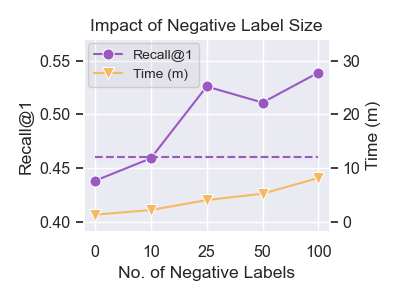}
    \caption{Impact of the number of negative classes on QueryAdapter performance and training time.}
    \end{subfigure}
\quad
    \begin{subfigure}[t]{0.3\textwidth}
    \centering
    \includegraphics[trim={0.3cm 0.5cm -0.5cm 0},clip,width=6cm]{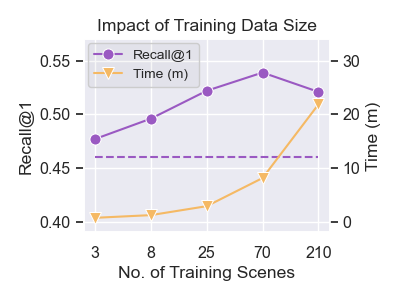}
    \caption{Impact of the number of scenes used for adaptation on QueryAdapter performance and training time.}
    \end{subfigure}
\caption{Impact of key parameters on QueryAdapter performance and training time using the small sets of target classes. The purple solid lines show the performance of the adapted model on the target classes. 
The purple dotted lines refers to performance of the pre-trained system on the target classes. 
The orange solid line shows the time taken to perform adaptation.}
\label{paramter_sensitivity}
\vspace{-1.5em}
\end{figure*}

\vspace{-0.3em}
\subsection{Task-oriented Object Retrieval}
We firstly assess the ability of our overall framework to respond to complex, task-oriented queries (Figure \ref{comparison_results}a). We compare our pipeline with ConceptGraphs \cite{conceptgraphs}, an approach for object retrieval from natural language based on 3DSGs. This approach incrementally merges object segments and features to produce a set of 3D objects. As in our approach, objects can be retrieved based on their cosine similarity with the query (CG Cosine). For more abstract queries, a caption is produced for each object and an LLM used to retrieve relevant objects (CG Captions). Owing to the inconsistent captions produced for each object, CG Captions performs poorly when responding to our task-based queries (Figure \ref{comparison_results}a). In particular, when relevant objects are missing from the 3DSG, the LLM is prone to producing an obscure plan in an attempt to respond to the query using the objects present. Using the LLM to first produce a set of target classes required to fulfil the query as per our approach avoids such confabulation. Using these target classes to query the 3DSG via cosine similarity thus produces a significant increase in average task recall (Figure \ref{comparison_results}a). However, this approach remains worse then querying the raw segments directly. This indicates that when merging object segments and features to produce a compact 3DSG, valuable information may be lost. Lastly, we see that QueryAdapter produces a further 6.7\% increase in average task recall on the task-oriented queries. This amounts to a 17.6\% increase in recall relative to using CG Captions. This result highlights the value of both QueryAdapter and our overall robotic vision paradigm in responding to complex natural language queries.

\vspace{-0.3em}
\subsection{Comparison of Unsupervised VLM Adapters}
We use the small sets of target classes to conduct a detailed comparison of unsupervised VLM adapters for query-oriented adaptation (Figure \ref{comparison_results}b). Due to the large number of open-query objects in the training set, UEO performs very poorly in this setting. UPL performs better, generating minor improvement relative to using the pre-trained model. 
% However, to make noticable difference to object retrieval, an alternative adaptation approach is required. In particular, new strategies are needed to deal with the many OOD objects present in the unlabelled data stream collected by the robot. 
However, QueryAdapter is the strongest approach, producing a 7.9\% improvement in recall@1 relative to the pre-trained model. This emphasises the importance of addressing the many open-query objects present in the raw data stream used for adaptation. 
% In practical terms, this result further shows that QueryAdapter can be used to improve the performance of a pre-trained VLM on a small set of object classes deemed relevant to the robot. 
Furthermore, it highlights that our negative labelling method, top $k$ object selection and the UEO loss are all integral to the effective operation of QueryAdapter.

\vspace{-0.3em}
\subsection{Ablation Study}
We also use the small sets of target classes to perform an ablation study of the proposed QueryAdapter (Figure \ref{comparison_results}c). We firstly assess the impact of running QueryAdapter without negative labels, which we term top $k$ only. 
% This approach performs well for the affordance queries, but actually harms performance for class queries relative to using the pre-trained model. 
This approach harms performance of the pre-trained model, emphasising the need for additional strategies to deal with open-query objects. In turn, the addition of our negative labelling approach is shown to improve recall@1 by 10.1\%. 
We also assess the inverse of this, where QueryAdapter is run with the negative labels but without top $k$ object selection. 
% Notably, this approach performs better for affordance queries than TaskAdapter. 
This approach is slightly worse than QueryAdapter, demonstrating that top $k$ object selection provides a small performance benefit in addition to improving efficiency. 
Lastly, we assess the impact of using random words as negative labels \cite{open_vlm}. This approach leads to a reduction of 5.2\% on recall@1 relative to QueryAdapter, further emphasising the value of using object captions as negative labels. 
% These results our negative labelling method, top $k$ object selection and the UEO loss are all integral to the effective operation of QueryAdapter. 

\vspace{-0.3em}
\subsection{Practical Considerations}
To minimise downtime of the robot, adaptation of the VLM to the current query needs to be performed as quickly as possible. Ultimately, QueryAdapter can produce an adapted VLM in a few minutes, which we argue is sufficient for many applications (Figure \ref{paramter_sensitivity}). However, there are several parameters that impact training time, such as top $k$, number of training scenes $j$ and number of negative labels $N$. A larger value for $k$ leads to more training samples being used, directly increasing training time (Figure \ref{paramter_sensitivity}a). However, this does not necessarily improve performance, as a larger $k$ introduces segments that are more likely to be OOD \cite{upl}. In practice, top $k=8$ generates optimal performance while maintaining low run-time. Secondly, the number of negative labels increases the time taken to calculate the UEO loss. There is a clear trade-off here, as using more negative labels tends to generate better performance (Figure \ref{paramter_sensitivity}b). We recommend using around 50 to 100 labels, as at this point performance appears to plateau. Lastly, we see that using more scenes for training increases both the training time and performance of QueryAdapter (Figure \ref{paramter_sensitivity}c). 
% However, even with only a few scenes, the adapted model performs better than the pre-trained VLM. 
This highlights the potential for QueryAdapter to be used in a practical continual learning setting, where as more scenes are explored by the robot its response to natural language queries will improve. 

\begin{table}[t]
\centering
\caption{Performance of QueryAdapter in alternative deployment scenarios. Small sets of target classes are used to evaluate performance on affordance-based queries in Scannet++. Additionally, the performance of the adapted VLMs produced for task-oriented queries in the Scannet++ dataset are evaluated on scenes from Ego4D. }
\begin{tabular}{l|l|l}
\hline
Method & Affordance  & Ego4D \\ \hline
QueryAdapter   & \textbf{30.84 (+10.64)} & \textbf{33.06 (+8.11)} \\ \hline

Pre-trained   &  20.19  & 24.95   \\ \hline
\end{tabular}
\vspace{-1.5em}
\label{tab:other}
\end{table}

\vspace{-0.3em}
\subsection{Alternative Deployment Scenarios}
We additionally explore the performance of QueryAdapter in alternative deployment settings (Table \ref{tab:other}). Firstly, we evaluate the potential for our method to improve performance on affordance queries. Such abstract queries are known to be challenging for CLIP-based retrieval systems, motivating the use of MLLMs in some work \cite{conceptgraphs}. To evaluate affordance queries, we perform adaptation using a new set of target classes where the object classes are replaced with common object affordances. These are generated by asking an LLM to define the most common use case for each class.
% Object retrieval is then evaluated for these affordance queries using the pre-trained and adapted model. 
In this setting, the adapted model generates an improvement in recall@1 of 10.6\%. Evidently, adaptation strategies such as QueryAdapter can be used to align CLIP features with more abstract concepts, potentially avoiding the need to use computationally expensive methods such as MLLMs.   

Lastly, we assess the ability of the adapted models to generalise across datasets and application domains. We take the adapted VLMs trained on the Scannet++ dataset and evaluate them on scenes from the Ego4D dataset \cite{ego4d, paco}. This dataset contains footage from wearable cameras showing people completing common manipulation tasks. Our evaluation procedure with this dataset remains the same as when using ScanNet++ to assess task-oriented queries. The only change is that we update the set of relevant classes for each query using the Ego4D labels, and a bounding-box Intersection over Union of 50 is used to associate ground-truth labels to predicted segments. Despite never having seen Ego4D data, the adapted model improves average task recall by 8.1\% in these scenes. This emphasises that QueryAdapter is robust to the visual appearance changes that can occur between datasets. Furthermore, this result highlights the potential for QueryAdapter to improve the execution of common manipulation tasks.

\vspace{-0.5em}
\subsection{Limitations and Future Work}
This work raises several directions for improving how VLMs are adapted for robotic deployment. Firstly, despite requiring only a few minutes to train, there may be opportunities to further improve the efficiency of QueryAdapter. For example, there may be solutions in the Test-Time Training (TTT) literature, which aims to perform adaptation online using a stream of images \cite{improved_tta}. However, how to perform query-oriented adaptation with such approaches remains unexplored. The incremental use of QueryAdapter could also be investigated in more detail. In particular, different adaptation strategies may be optimal in low data scenarios \cite{coop, cocoop} in comparison to when training data is plentiful \cite{udaod2}. Lastly, the robotic vision framework proposed in this paper is yet to be integrated with downstream methods of task execution. This process could have interesting implications for how open-vocabulary robotic vision systems are evaluated. For example, it is unclear how existing open-vocabulary systems would respond if a queried object is not present in the scene.

\vspace{-0.3em}
\section{Conclusion}
In this paper, we aim to adapt a pre-trained VLM for use in robotic deployment environments \textit{without} pre-defining a closed-set of classes. To this end, QueryAdapter is explored to rapidly adapt a pre-trained VLM in response to natural language queries. Concurrently, a method is proposed to perform adaptation using a raw data stream containing many open-query objects. This approach has been demonstrated to improve object retrieval from natural language queries in a variety of real-world scenes. We anticipate that this research will guide how VLMs should be adapted for use in downstream robotic tasks.

\bibliographystyle{IEEEtran}
\bibliography{IEEEabrv,main}

\end{document}